\documentclass[conference]{IEEEtran}
\IEEEoverridecommandlockouts
\usepackage{cite}
\usepackage{amsmath,amssymb,amsfonts}
\usepackage{graphicx}
\usepackage{textcomp}
\usepackage{url}
\usepackage{balance}
\usepackage[table,xcdraw]{xcolor}
\usepackage[inline]{enumitem}
\usepackage{booktabs}
\usepackage[ruled,linesnumbered]{algorithm2e}
\usepackage{svg}

\newcolumntype{L}[1]{>{\raggedright\let\newline\\\arraybackslash\hspace{0pt}}m{#1}}
\newcolumntype{C}[1]{>{\centering\let\newline\\\arraybackslash\hspace{0pt}}m{#1}}
\newcolumntype{R}[1]{>{\raggedleft\let\newline\\\arraybackslash\hspace{0pt}}m{#1}}

\RestyleAlgo{ruled}
\SetKwComment{Comment}{// }{}

\newcommand{\q}[1]{``#1''}

\def\BibTeX{{\rm B\kern-.05em{\sc i\kern-.025em b}\kern-.08em
    T\kern-.1667em\lower.7ex\hbox{E}\kern-.125emX}}
\begin{document}

\title{Detecting Relevant Information in High-Volume Chat Logs: Keyphrase Extraction for Grooming and Drug Dealing Forensic Analysis\\

\thanks{J. H. Alves received funding by the \textit{Coordenação de Aperfeiçoamento de Pessoal de Nível Superior (CAPES)} via the \textit{Programa Nacional de Cooperação Acadêmica} (PROCAD-SPFC) program.}
}

\author{
\IEEEauthorblockN{Jeovane Honório Alves\IEEEauthorrefmark{1}, Horácio A. C. G. Pedroso\IEEEauthorrefmark{1}, Rafael Honorio Venetikides\IEEEauthorrefmark{3}, \\Joel E. M. Köster\IEEEauthorrefmark{4}, Luiz Rodrigo Grochocki\IEEEauthorrefmark{4}, Cinthia O. A. Freitas\IEEEauthorrefmark{2}, Jean Paul Barddal\IEEEauthorrefmark{1}}
\IEEEauthorblockA{\IEEEauthorrefmark{1}Programa de Pós-Graduação em Informática (PPGIa), Pontifícia Universidade Católica do Paraná (PUCPR), Curitiba, Brazil}
\IEEEauthorblockA{\IEEEauthorrefmark{2}Programa de Pós-Graduação em Direito (PPGD), Pontifícia Universidade Católica do Paraná (PUCPR), Curitiba, Brazil}
\IEEEauthorblockA{\IEEEauthorrefmark{3}Escola Politécnica, Pontifícia Universidade Católica do Paraná (PUCPR), Curitiba, Brazil}
\IEEEauthorblockA{
\IEEEauthorrefmark{4}Polícia Científica do Paraná (PCPR), Curitiba, Brazil\\
}
Email: jeovane.honorio@pucpr.br, 
horacio.pedroso@pucpr.edu.br,
rafael.venetikides@pucpr.edu.br, \\
joel.koster@policiacientifica.pr.gov.br,
luiz.grochocki@policiacientifica.pr.gov.br,\\
cinthia.freitas@pucpr.br, jean.barddal@ppgia.pucpr.br
}

\maketitle

\begin{abstract}
The growing use of digital communication platforms has given rise to various criminal activities, such as grooming and drug dealing, which pose significant challenges to law enforcement and forensic experts. This paper presents a supervised keyphrase extraction approach to detect relevant information in high-volume chat logs involving grooming and drug dealing for forensic analysis. The proposed method, JointKPE++, builds upon the JointKPE keyphrase extractor by employing improvements to handle longer texts effectively. We evaluate JointKPE++ using BERT-based pre-trained models on grooming and drug dealing datasets, including BERT, RoBERTa, SpanBERT, and BERTimbau. The results show significant improvements over traditional approaches and demonstrate the potential for JointKPE++ to aid forensic experts in efficiently detecting keyphrases related to criminal activities.
\end{abstract}


\section{Introduction}

The increasing usage of the Internet and the widespread availability of smartphones have led to a significant rise in messaging exchanges. 
As smartphones become more accessible to a broader range of users, messaging has become a prevalent mode of communication.
However, along with the growth in mobile device usage, there has also been an increase in criminal activities involving the exchange of illicit messages through messaging apps and SMS. 
These messages can contain various forms of criminal behavior, such as threats, defamation, explicit content, or even discussions related to drug dealing. To ensure the safety and protection of potential victims, effectively identifying and investigating these criminal activities is of utmost importance.

In this context, forensic analysis for smartphones and messaging apps plays a vital role in investigating criminal messages. Forensic experts specialize in extracting data from these devices and conducting thorough analyses to determine the origin and authenticity of the messages, as well as gather other relevant information crucial to the investigation. The extracted data can serve as valuable evidence in legal proceedings, aiding in the conviction of the individuals responsible for these criminal acts \cite{martinez2020information}.

Computational tools are vital to support forensic experts during analyses as they expedite the process while enhancing accuracy and automating several steps of data acquisition and analysis from smartphones and other devices. These tools efficiently extract relevant data and information from various data sources, including text messages, images, audio, and more, streamlining the analysis of vast data volumes and reducing manual effort.

Within extensive text volumes, specific keyphrases often contain valuable content. Recognizing this, using keyphrase extraction techniques proves invaluable for forensic investigators in detecting criminal information within extensive text content found in chats. While investigators can perform simple keyword searches using text editors, criminals frequently employ coded language or other strategies to impede forensic detection. Therefore, employing robust and context-based automatic keyphrase extractors significantly contributes to the forensic analysis of diverse types of crimes.

This work introduces a supervised keyphrase extraction approach for long-context texts to advance forensic research in online chats involving grooming to practice sexual abuse and drug dealing. For the first case, a publicly available subset from the Perverted Justice data was experimented with. As for the second case, we used private data obtained through a partnership with the Scientific Police of Paraná (\textit{Polícia Científica do Paraná} -- PCPR), a local police force from Brazil.

The proposed method builds upon a modified version of JointKPE, a supervised keyphrase extraction technique. We enhanced it to handle high-volume chat logs, thus creating JointKPE++ to aid forensic experts in efficiently detecting crucial information within a large volume of messages from messaging apps and SMS. This enhancement represents a step forward in research in this area, promising improved results for keyphrase extraction tasks in digital forensic analysis.

This paper is structured as follows: Section \ref{sec:related} introduces the study cases of grooming and drug dealing, along with related works. Section \ref{sec:proposed} presents our proposed approach. Section \ref{sec:experiments} discusses experimental settings and results. The paper concludes with remarks in Section \ref{sec:conclusion}.
\section{Criminal Activities in Chats} 
\label{sec:related}

In our interconnected world, criminal activities in online chats, particularly on smartphones, have become increasingly prevalent. 
Research on online chats faces limitations in accessing real enticement or drug dealing cases due to the confidential nature of legal proceedings, particularly concerning the involvement of children or teens. These cases are subject to judicial secrecy to protect the privacy of individuals and prevent the disclosure of confidential information such as wiretaps and bank statements. As a result, the availability of databases containing such cases is limited.

In this section, we discuss the specific criminal activities examined in this study, the data used, and the challenges associated with automating the detection of relevant information in online chats. The focus is on grooming and drug dealing conversations.

\subsection{Child Grooming}

The use of online chats and smartphones for grooming is a serious concern requiring the attention and collaboration of parents, authorities, and society. Sexual predators exploit the anonymity and convenience of digital chats to groom vulnerable children and teens, who may be unaware of the risks involved, and face social and family challenges. These individuals employ manipulative tactics, such as desensitization techniques, to engage victims in sexual abuse.
In 2022, the National Center for Missing and Exploited Children (NCMEC) received over 80,000 reports of online grooming of children for sexual acts, representing a significant increase from the previous year's reports (approximately 44,000 in 2021) \cite{NCMEC22}. These heinous acts can have lasting effects on victims and their families while enabling criminals to continue their abuse.

To address this critical issue, several studies have utilized chat logs from the Perverted Justice (PeeJ) project to develop studies for detection of content related to grooming and sexual abuse of children and teens in conversations, especially chat log classification (i.e., if a chat contains evidence of some grooming). PeeJ is an online initiative that exposes individuals involved in grooming and sexually abusing children or teens. Adult volunteers pose as children/teens in online conversations to identify potential predators \cite{egan2011perverted}.

These works mostly use a subset of the PeeJ chat logs and other non-predatory ones, creating unique datasets. ChatCoder was proposed in \cite{kontostathis2009chatcoder} to classify user communication between predator and victim and whether a chat log has grooming content. The authors grouped sentences into eight stages based on keyphrases found in them.  
These stages are based on the Luring Communication Theory, introduced in \cite{olson2007entrapping}, which identifies the stages an aggressor goes through to attract victims in the real world. This theory has been adapted to the digital realm in \cite{leatherman2009luring}, enabling the characterization of the stages an aggressor engages in through chat room conversations.

In the PAN-2012 sexual predator identification competition \cite{inches2012overview}, two challenges were conducted: grooming conversation and grooming sentence classification. Competitors utilized lexical and behavioral features with machine learning classifiers. For the first challenge, F1-scores up to 87\% were achieved, while the second challenge had F1-scores up to 30\%. Notably, the absence of pre-existing ground-truth for the second challenge impacted the evaluation, potentially inflating the overall results.
An updated version of ChatCoder \cite{kontostathis2012identifying} was employed for detecting groomers and grooming sentences using the dataset from the PAN-2012 competition. They used decision trees and rule sets, achieving an F1 score of 39\% for grooming sentence classification.

In \cite{vogt2021early}, the emphasis is on early detection of predatory messages, explicitly analyzing the chat from its inception until the end, aiming to detect the initial attempts at grooming as early as possible. They propose a two-tier approach that utilizes BERT for analyzing message windows in a simulated ongoing chat and classifying window sequences.
A hybrid sampling approach was used in \cite{borj2021detecting} to address class imbalance in grooming detection. Combining class re-distribution, data augmentation techniques, and a Histogram Boosted Gradient classifier, an F1-score of 99\% was achieved in the PAN-2012 dataset.
A contrastive learning approach was proposed in \cite{rezaee2023detecting} for classifying grooming conversations using BERT-based model embeddings. The fusion-based classifier achieved an F1-score of 97\%.

A survey of sexual grooming was conducted in \cite{borj2022online}. Various features such as bag of words, word embedding, affective-based (e.g., Linguistic Inquiry and Word Count -- LIWD), statistical chat-based, and keystroke dynamics-based features were employed in grooming analysis. Traditional machine learning classifiers (e.g., SVM, kNN, Random Forest, XGBoost) and deep learning models (e.g., CNN, BiLSTM) were commonly used for grooming conversation, stage, and sentence classification tasks. F1-scores for grooming detection reached up to 96\% in different datasets and 87\% for the PAN-2012 dataset. While grooming detection showed promising results overall, further experiments on variable and more realistic data are necessary. Additionally, research in other tasks is crucial for improving grooming detection from different perspectives.

Most research on grooming detection emphasizes conversation classification, demonstrating promising results despite limited data. Sentence classification, primarily observed in the PAN-2012 challenge, has room for improvement. To the best of our knowledge, there is a lack of robust keyphrase extraction techniques applied to grooming chat logs. Exploring keyphrase extraction in grooming analysis can be valuable for forensic experts to enhance detection and gain a deeper understanding of grooming techniques, facilitating the presentation of evidence in court.

\subsection{Drug Dealing}

Drug dealing has become increasingly prevalent in social media and chat applications, such as WhatsApp, as criminals leverage these platforms to facilitate their illicit trade. They establish connections with other criminals and potential clients, expanding their relationship networks while employing coded and discreet language to evade detection or dissimulate the investigation \cite{demant2019drug}.

When smartphones associated with drug dealing are seized, forensic teams often encounter a substantial volume of chat logs, as these devices are commonly used for personal communication as well. Analyzing the entirety of these chat logs is a significant challenge. Even when employing a predefined list of keywords for a keyword-based search, relevant information is not guaranteed to be uncovered. This is primarily because criminals utilize a distinct coded language that may differ from the selected keywords, making it difficult to identify crucial details through conventional search approaches.

Several studies have utilized machine learning techniques to distinguish between texts with and without drug dealing content.
In \cite{yang2017tracking}, a multimodal approach combining image and text data was employed to classify Instagram posts. The study utilized a decision-level fusion method to integrate classifiers based on both images (GoogLeNet) and text (n-grams and TF-IDF). Data was collected from Instagram and Google to create the datasets.
Mackey et al. \cite{mackey2018solution} used the unsupervised Biterm Topic Model (BTM) to extract text patterns and summarize content into topics for detecting illicit drug dealings on Twitter.

Li et al. \cite{li2019machine} compared traditional machine learning models and an LSTM approach for classifying Instagram posts on illicit drug dealing. Using only text data (i.e., without hashtags) achieved better results, with the LSTM approach outperforming traditional models. In their subsequent work \cite{shah2022unsupervised}, they employed BTM. The study examined both parent posts and comments, revealing the presence of illicit content in comments, even when the parent post was unrelated to drug dealing.

A multimodal approach using heterogeneous graphs (HG) and relation-based graph convolutional networks (R-GCNs) is proposed in \cite{qian2021distilling} for analyzing Instagram data. Graph structure refinement (GSR) and meta-learning are employed to enhance node representations and address limited labeled data, resulting in improved performance.
A multimodal approach for Instagram is utilized in \cite{hu2021identifying}, combining a BERT model for text classification and a ResNet model for image classification. This approach focuses on analyzing content in profiles and posts.

Many studies primarily concentrate on public data from platforms like Twitter and Instagram, utilizing web scraping techniques at various time intervals. However, in forensic analysis, the focus shifts to extracting and examining data from private messaging apps such as WhatsApp. There is a need for the development of approaches that can effectively analyze the vast volume of text messages from these messaging apps to extract relevant information.
\section{Proposed Approach}
\label{sec:proposed}

Given chats' nature, many messages may be unrelated to the specific context of interest. 
For instance, personal conversations or exchanges with adults are likely to be irrelevant to drug dealing or grooming conversations, respectively (although they may contain indirect information about such crimes). 
As a result, most chat logs within a person's smartphone does not pertain to contents related to criminal activities. 
Furthermore, there may be a high volume of text in these chats. 
We employ a supervised keyphrase extraction approach for high-volume chat logs to address these challenges.

The purpose of keyphrase extraction is to efficiently and succinctly extract relevant content of a text. 
These techniques can be classified into different approaches \cite{siddiqi2015keyword}. 
Unsupervised techniques, such as TF-IDF \cite{ramos2003using}, calculate term importance based on frequency within a document and the corpus. 
RAKE \cite{rose2010automatic} determines word relevance based on co-occurrence ratios. 
TextRank \cite{mihalcea2004textrank} uses a graph structure to assess word strength and similarity. 
Unlike the above techniques, KeyBERT \cite{grootendorst2020keybert} employs supervised learning, using pre-trained BERT embeddings \cite{devlin2018bert} and cosine similarity for importance and relatedness determination.

While KeyBERT utilizes pre-trained BERT embeddings, it does not employ finetuning. 
However, in keyphrase extraction for chat logs containing criminal content, unsupervised and supervised methods without finetuning may yield sub-optimal results as the extracted keyphrases could include irrelevant parts of the chat log, such as personal or professional conversations that are legal. 
We employ a supervised method with finetuning called JointKPE to overcome this limitation. 

JointKPE, introduced in \cite{sun2021capturing}, is a supervised keyphrase extraction approach that utilizes BERT-based pre-trained models as its encoder. 
It employs two strategies, informative ranking and keyphrase chunking, to enhance the informativeness and phraseness of the extracted keyphrases. 
The BERT model encodes a text document into a sequence of word embeddings, which is then inputted to convolutional modules to generate an n-gram representation. 
Global informativeness scores are computed using a linear classifier and Margin Ranking loss. 
Additionally, a keyphrase chunking task is performed, optimizing a binary classifier to match the n-gram representation with the annotated keyphrases in the ground truth. The loss functions of both ranking and chunking tasks are combined to achieve a balanced optimization between the two objectives.

The BERT encoder's maximum document length of 512 tokens (including \q{[CLS]} and \q{[SEP]} special tokens) poses challenges for evaluating high-volume chat logs, some of which exceed 50 thousand words. The vanilla JointKPE approach truncates the text after 510 tokens (since two tokens are used by the special tokens, reaching 512 tokens in total), extracting only keyphrases for chat logs with smaller conversations, which may represent a minority of the samples (or only extracting keyphrases from the beginning of the text). 
One possible solution to alleviate this limitation is to split the chat logs into blocks of 512 tokens and then combine the keyphrase candidates predicted by the approach. 
However, this method still evaluates candidates within local text blocks, which might restrict the scope of JointKPE's analysis.
  
\begin{figure}[ht!] 
\centering
\includegraphics[width=0.8\linewidth]{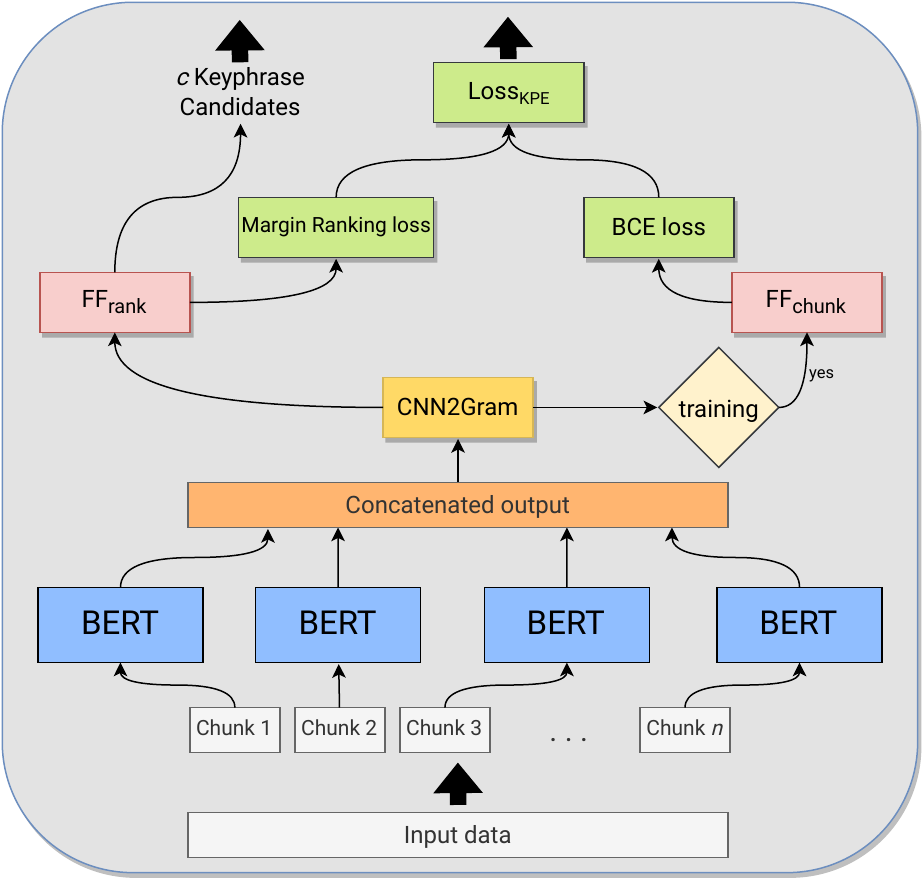}
\caption{Overall workflow of the proposed JointKPE++ approach.}
\label{fig:workflow}
\end{figure}

To expand the scope of JointKPE, we enhance the approach by enabling the BERT encoder to process multiple text blocks in a single execution, extending the document length to 8192 tokens (including special tokens). 
The approach involves splitting a large portion of text into text blocks of 512 tokens and inputting each to the BERT encoder. 
After processing the text blocks with BERT, we concatenate them into a single sequence output, which serves as the input for a convolutional layer. 
This augmentation increases the scope evaluated by the JointKPE method. 
We can increase the inputted number of tokens without excessive memory consumption by employing different code optimizations, like automatic mixed precision (AMP). 
For samples with more than 8192 tokens, we can split them during both the training and evaluation stages. 
During the evaluation stage, we join the keyphrase candidates and rank them based on the outputted score.
Figure \ref{fig:workflow} shows the overall workflow of our enhanced JointKPE.

Algorithm \ref{alg:jointKPE++} provides a pseudo-code description of the enhanced JointKPE approach, conveniently referred to as JointKPE++. This algorithm exhibits two main parts, with the first part (Lines 1--8) featuring modifications in comparison to the original approach, while the second part remains consistent with JointKPE.

In the initial stage, JointKPE++ takes as input a document text, denoted as $D_N$, comprising a total of $N$ tokens. In Line 3, $D_N$ undergoes a segmentation into $n$ blocks, each consisting of $m$ tokens, thereby producing $B_n$. Following this, in Lines 4-5, each block $B_i$ is encoded using a BERT encoder, resulting in the creation of subsequence $subseq_i$. These generated subsequences, represented as $subseq_i^n$, are concatenated to yield $seq$ (Line 7). This concatenated sequence is subsequently fed into a sequence of convolutional modules (Line 8). Each module is equipped with a 1D convolutional layer having a kernel size $i$, used to generate scores for various combinations of keyphrases with different n-gram sizes ($1 \le i \le k$). In our experiments, we set the maximum n-gram size as $k=7$. This sequence of convolutional modules is collectively referred to as the CNN2Gram function. The outcomes of these modules are then concatenated to produce $out_{cnn}$.

In the second stage, the $out_{cnn}$ serves as input for the informative ranking classifier (Line 9), which consists of a feed-forward layer responsible for generating ranking scores $scores_{rank}$ for the keyphrase candidates. These scores are employed in Lines 11-12 to compute the ranking loss $loss_{rank}$. It is noteworthy that each keyphrase candidate can appear in different locations and contexts within the text, leading to diverse localized informativeness scores. A max function is used to obtain global informativeness scores for each keyphrase candidate. Then, in Line 12, a Margin Ranking Loss is applied to the global scores, as defined by the function $\sum max(-global^+ + global^- + 1)$.

During the training phase exclusively (Lines 13-15), the chunking loss is determined using another feed-forward layer designed for the chunking task (Line 13). A binary cross-entropy (BCE) loss is applied to its output. Subsequently, the chunking loss is combined with the ranking loss to yield $loss_{kpe}$ -- the overall loss.
Finally, in Line 16, the keyphrase candidates are obtained, returning the $c$ best keyphrase candidates ranked by their scores (from $scores_{rank}$).

\begin{algorithm}[!t]
\caption{Algorithm of the JointKPE++ approach to support longer texts}\label{alg:jointKPE++}
\KwData{$D^N$ \Comment*[r]{Text with $N$ tokens}}
\KwResult{$K^c$, $\textrm{loss}_{\textrm{kpe}}$}
$m \gets 512 $\;
$k \gets 7$\;
$B^n \gets \textrm{split}(D^N, m)$\;

\For{i=1; i $\le$ n; i++}{
    $\textrm{subseq}_i \gets \textrm{BERT}(B_i)$\;  
}

$\textrm{seq} \gets \textrm{concat}(\textrm{subseq}_i^n)$\;  
$\textrm{out}_{cnn} \gets \textrm{concat}(\textrm{cnn2gram}_i^k(\textrm{seq}))$\; 
$\textrm{scores}_{\textrm{rank}} \gets \textrm{feed\_forward}_{\textrm{rank}}(\textrm{out}_{\textrm{cnn}})$\; 
\If{\textrm{training}}{
    $\textrm{global}_{\textrm{rank}} \gets \max(\textrm{scores}_{\textrm{rank}})$\; 
    $\textrm{loss}_{\textrm{rank}} \gets \textrm{margin\_loss}(\textrm{global}_{\textrm{rank}}^{+}, \textrm{global}_{\textrm{rank}}^{-})$\; 
    $\textrm{scores}_{\textrm{chunk}} \gets \textrm{feed\_forward}_{\textrm{chunk}}(\textrm{out}_{\textrm{cnn}})$\;  
    $\textrm{loss}_{\textrm{chunk}} \gets \textrm{bce}\_\textrm{loss}_{\textrm{chunk}}(\textrm{scores}_{\textrm{chunk}})$\;  
    $\textrm{loss}_{\textrm{kpe}} \gets \textrm{loss}_{\textrm{rank}} + \textrm{loss}_{\textrm{chunk}} $
}

$K^c \gets \textrm{get\_candidates}(\textrm{scores}_{\textrm{rank}}, c)$ \Comment*[r]{$c$ keyphrase candidates}
\end{algorithm}

\section{Experiments} \label{sec:experiments}

This section describes the experiments conducted to assess JointKPE++ for keyphrase extraction in grooming and drug-dealing scenarios.
First, we detail the datasets and the experimental settings used.
Next, we describe and discuss the outcomes observed.

\subsection{Datasets}

In this section, the datasets used in our experiments are described. We used a publicly available dataset (in English) for the grooming study case and a private dataset (in Portuguese) for the drug dealing study.

\subsubsection{Text Mining and Cybercrime dataset}

In our research on grooming chat logs, we used a dataset developed in \cite{kontostathis2010text}. The TMC (Text Mining and Cybercrime -- we used the paper's name to specify the chat logs and annotations employed in our work) dataset contains 288 chat logs in English from the Perverted Justice website and is publicly available. 
One thing to note is that only the sentences texted by the groomers were annotated. However, the entire chat log is processed in our experiments,  thus simulating a real-life forensic analysis. 

The authors annotated keyphrases based on eight stages of grooming (previously applied in \cite{kontostathis2009chatcoder}).
However, our proposed approach does not make distinctions between these stages, as our primary focus is on keyphrase extraction rather than their classification.
We excluded seven chat logs as some were partially annotated, and others had no annotation. For the remaining 281 chat logs, we formatted the data using only the message content in sequential order for our experiments. The chat logs varied from 274 to 73,040 words, averaging 5,683 words per chat log. Of the 281 chat logs, 271 (97.15\%) contained more than 512 words.

The keyphrases were originally available after a preprocessing. For example, the keyphrase \textit{"i want u"} was available as \textit{"i want you"}, which, in some cases, the preprocessed keyphrase was absent in the chat log. Thus, we applied several rule-based techniques to obtain the keyphrases in the original data. Table \ref{tab:tmc_examples} presents a comparison between the formatted and original text for some sequence examples from the TMC dataset. It's worth noting that the formatted text, in certain instances, included only the root word as the keyphrase and even different words compared to the original text.

\begin{table}[htb]
\centering
\caption{Comparison between the formatted and original text from the TMC dataset. Keyphrases are in \textbf{bold}.}
\label{tab:tmc_examples}
\begin{tabular}{R{4cm}|L{4cm}}
\toprule
Formatted text & Original text \\
\toprule
than you should go get some sleep \textbf{sex}y & than u should go get some sleep \textbf{sexy} \\
i just remembered which \textbf{pictures} you were talking about & I just remembered which \textbf{pics} you were talking about \\
how big are your \textbf{nipples} & how big are your \textbf{tits} \\
i'd love to be \textbf{suck}ing your \textbf{nips} & i'd love to be \textbf{sucking} your \textbf{nips} \\
okay maybe a button down shirt unbuttoned no \textbf{brassiere} & k maybe a button down shirt unbuttoned no \textbf{bra} \\
\textbf{what do you want to do} & \textbf{what do u wanna do}  \\
\bottomrule
\end{tabular}
\end{table}

\subsubsection{Drug dealing dataset}

The drug dealing dataset is a private dataset containing ten forensic cases, where each case contains several WhatsApp conversations in Portuguese extracted from a seized smartphone. 

The chat log data from these cases was obtained with the cooperation of PCPR investigators and are related to court cases and forensic investigations. Thus, it is impossible to publish the dataset or present significant data from the chat logs. However, evaluating our approach in this data is extremely important since we are testing on real data evaluated by forensic experts.

The ground-truth keyphrases underwent validation by PCPR investigators, ensuring the dataset's reliability and the accuracy of its annotated keyphrases.
Of all the cases, only two had less than two thousand words, while the others contained 21 thousand to over one million words. On average, each case consisted of 218,393 words.

\subsection{Experimental Settings}

We use the AdamW optimizer with the one-cycle cosine annealing learning rate (LR) scheduler for training. We start with an initial warm-up and then continue with LR decrement after reaching the value of 5e-5. Given that the datasets contain ground-truth keyphrases with more than five words, we set the maximum phrase length to $k=7$. 

Our training is performed online (batch size $=1$) on an RTX 3080 with 24GB, where we employ AMP to reduce memory consumption. The document text has a maximum number of tokens $N=8192$ (including special tokens) for the grooming experiments. As the private drug dealing dataset data must remain within the police force servers, we conduct experiments on an NVIDIA Quatro RTX 4000 with 8GB and $N=3072$.
For the grooming experiments, we trained JointKPE++ for 50 epochs. We evaluated it using three BERT-based pre-trained models: vanilla BERT, RoBERTa, and SpanBERT (employed in the JointKPE paper). However, for the drug dealing experiments, we only employed the BERTimbau model \cite{souza2020bertimbau} due to the scarcity of RoBERTa and SpanBERT pre-trained models for Portuguese documents, training it for 20 epochs.

For evaluation, we utilize the F1-score to assess the performance of the $K$ extracted keyphrases (F1@$K$).
The predicted keyphrases are considered correct only if they exactly match a keyphrase from the ground-truth list. 
In contrast to prior studies such as \cite{meng2017deep, xiong2019open, sun2021capturing}, where $K$ values of $10$ or less were used due to the relatively limited number of keyphrases within their datasets, we have opted for different $K$ values in our research. 
Specifically, for the TMC dataset, we have selected $K=\{40,50,60\}$, while for the drug dealing dataset, we have chosen $K=\{20,30,40\}$. 
This decision aligns with the observation that both of our evaluated datasets exhibit a notably higher prevalence of keyphrases.
We utilize a five-fold cross-validation approach for both datasets, ensuring a balanced distribution of words across the folds.

\subsection{Grooming Keyphrase Extraction}

We comprehensively compare our proposed approach with some traditional keyphrase extractors, including TF-IDF, RAKE, and TextRank. Additionally, we compare our method against KeyBERT, which utilizes MPNetV2 as its encoder.
We present results from various scenarios to evaluate the effectiveness of our approach compared to the original JointKPE method. Firstly, we show the performance of the vanilla JointKPE approach, where documents are trimmed during training and testing to fit the maximum token limit (i.e., 512 tokens). Secondly, we explore the performance of JointKPE with a joined evaluation, where training documents are trimmed, and validation documents are split, processed by JointKPE, and then evaluated together. Lastly, we examine the results of the joined JointKPE approach, where training documents are split and trained separately, while validation documents are evaluated jointly (like the second scenario).

By conducting these comparisons, we aim to demonstrate the superiority of our proposed approach over traditional keyphrase extractors and its potential enhancements over the original JointKPE method. Table \ref{tab:grooming_results} shows the overall results obtained in the TMC dataset, including three different pre-trained models evaluated with our JointKPE++ approach.

\begin{table}[htb]
\centering
\caption{Overall F1-scores from different approaches of keyphrase extraction applied to the TMC dataset (in \%).}
\label{tab:grooming_results}
\begin{tabular}{lrrr}
\toprule
Approach & F1@40 & F1@50 & F1@60 \\ \midrule
TF-IDF & 2.89 & 3.46 & 3.99 \\
RAKE & 1.22 & 1.48 & 1.63 \\
TextRank & 5.15 & 6.02 & 6.59 \\
KeyBERT & 13.77 & 14.01 & 14.10 \\ \midrule
Vanilla BERT-JointKPE & 24.70 & 22.01 & 19.91 \\
Joined-eval only BERT-JointKPE & 46.73 & 44.70 & 42.68 \\ 
Joined BERT-JointKPE & 52.35 & 50.68 & 48.17 \\ \midrule
BERT-JointKPE++ & 66.42 & 67.15 & 65.71 \\ 
SpanBERT-JointKPE++ & \textbf{67.52} & \textbf{68.03} & \textbf{66.55}  \\
RoBERTa-JointKPE++ & 58.32 & 57.93 & 56.13 \\ \bottomrule
\end{tabular}
\end{table}

The evaluations shed light on the efficacy of our method in handling longer documents and its ability to provide more accurate and informative keyphrase predictions in real-world scenarios. 
The traditional keyphrase extractors showed sub-optimal results, achieving F1 scores below 10\%. KeyBERT demonstrated better performance with an F1@50 of 14\% but still lacked fine-tuning, leading to irrelevant keyphrase extraction for forensic analysis.
Moving on to the evaluation of the vanilla BERT-JointKPE, we observed improvements even with data trimming, achieving an F1@50 of 22\%. The proposed scenarios further enhanced the results. By splitting chat logs during validation and evaluating the split text blocks together, we achieved an F1@50 of 44.70\%. The default approach of trimming the remaining text after 512 tokens proved unsuitable, emphasizing the need for alternative strategies. In the last scenario, we split the training chat logs to employ the entire text data for training, while the validation stage remained similar to the previous scenario. This change in the training approach resulted in significant improvement, achieving the best F1@50 of 50.68\% for the JointKPE approach.

Our proposed method, JointKPE++, performed better than the evaluated approaches. As a direct enhancement of the original JointKPE, our approach achieved an F1@50 of 67.15\% with BERT as its encoder. We also employed different pre-trained models, SpanBERT and RoBERTa. The best results were obtained with SpanBERT, achieving an F1@50 of 68\%, while Roberta had inferior results, with an F1@50 of 57.93\%.

\subsection{Drug Dealing Keyphrase Extraction}

In Table \ref{tab:drug_dealing_results}, we present the results obtained in the private dataset from the police force. We employ experiments in traditional keyphrase extractors, KeyBERT with MPNetV2 (instead of BERTimbau since it had superior results), and JointKPE++ using BERTimbau as its pre-trained model. Unreported results occurred when the recall was zero, indicating that no ground-truth keyphrase was extracted, resulting in F1 scores equal to zero.

\begin{table}[htb]
\centering
\caption{Overall F1-scores from different approaches of keyphrase extraction applied to the private drug dealing dataset (in \%).}
\label{tab:drug_dealing_results}
\begin{tabular}{lrrr}
\toprule
Approach & F1@20 & F1@30 & F1@40 \\ \midrule
TF-IDF & 0.38 & 0.32 & 0.27 \\
RAKE & - & 0.57 & 0.44 \\
TextRank & - & - & 0.45 \\
KeyBERT & 0.80 & 0.57 & 0.44 \\ \midrule
BERTimbau-JointKPE++ & \textbf{23.09} & \textbf{22.53} & \textbf{20.51} \\ \bottomrule
\end{tabular}
\end{table}

The dataset used in this study contained even more words per case compared to the TMC dataset (average of 218,393 words against 5,683 words), which significantly increased the complexity of the task, particularly for traditional methods. As observed, these methods achieved F1 scores below 1\%, and in some scenarios, no ground-truth keyphrase was extracted. In contrast, our approach showed promising results, achieving an F1@30 of 22.53\%.

In conclusion, the modifications employed in JointKPE++ led to significant improvements, demonstrating the importance of fine-tuning and supporting longer texts in these types of problems.
\section{Conclusion} \label{sec:conclusion}

Detecting relevant information for forensic analysis in high-volume chat logs can be challenging. Criminals often use coded language, slang, and typos, complicating manual searches and reducing the ability to identify all criminal content. Traditional keyphrase extractors focus on keyphrases representing the entire chat log, which may not capture criminal-related messages, which are the minority of cases.

To address this issue, we use JointKPE, a supervised keyphrase extractor fine-tuned for better adaptation to criminal content. We also modify this approach to support long-context texts, which we call JointKPE++. Our approach outperforms traditional approaches, KeyBERT, and the vanilla JointKPE, showing promising results. JointKPE++ can significantly contribute to advancing research in detecting keyphrases related to criminal activities, enabling more efficient forensic analysis and reducing workload.

In JointKPE++, we utilize BERT in $n$ chunks of 512 tokens, processing each chunk separately, and then concatenate the outputs before inputting them into a 1-dimensional convolutional layer. However, the main limitation lies in the input size of the convolutional layer. Because of memory constraints, the input data must had at most 8192 tokens (i.e., $N \le 8192$). For longer texts, this is achieved by dividing them into balanced samples, each with $N \le 8192$ tokens. Each sample is processed separately, even during validation. However, during evaluation, the candidates from every sample (from the original input data) are considered jointly, with candidate duplicates resolved by selecting the best score.

While this approach allows for the evaluation of significantly longer texts (ideally speaking, an infinite amount of text), there remains a limit in terms of sequence length, capped at 8192 tokens (also, BERT is capped at 512 tokens), that can be processed jointly. For further research, it is essential to explore strategies to increase or eliminate these limits entirely. This would enable more reliable keyphrase extraction by considering the entire context together, extracting pertinent information based on the entirety of the text, rather than on separate blocks. 

Furthermore, expanding experiments to include more diverse data is encouraged, incorporating a more comprehensive range of chat logs from the two study cases and other criminal activities. By doing so, keyphrase extraction research can be more comprehensively evaluated and matured, ultimately contributing to its effective utilization by forensic experts.

\balance
\bibliographystyle{IEEEtran}
\bibliography{IEEEabrv,biblio.bib}

% Generated by IEEEtran.bst, version: 1.14 (2015/08/26)
\begin{thebibliography}{10}
\providecommand{\url}[1]{#1}
\csname url@samestyle\endcsname
\providecommand{\newblock}{\relax}
\providecommand{\bibinfo}[2]{#2}
\providecommand{\BIBentrySTDinterwordspacing}{\spaceskip=0pt\relax}
\providecommand{\BIBentryALTinterwordstretchfactor}{4}
\providecommand{\BIBentryALTinterwordspacing}{\spaceskip=\fontdimen2\font plus
\BIBentryALTinterwordstretchfactor\fontdimen3\font minus \fontdimen4\font\relax}
\providecommand{\BIBforeignlanguage}[2]{{%
\expandafter\ifx\csname l@#1\endcsname\relax
\typeout{** WARNING: IEEEtran.bst: No hyphenation pattern has been}%
\typeout{** loaded for the language `#1'. Using the pattern for}%
\typeout{** the default language instead.}%
\else
\language=\csname l@#1\endcsname
\fi
#2}}
\providecommand{\BIBdecl}{\relax}
\BIBdecl

\bibitem{martinez2020information}
J.~L. Martinez-Rodriguez, A.~Hogan, and I.~Lopez-Arevalo, ``Information extraction meets the semantic web: a survey,'' \emph{Semantic Web}, vol.~11, no.~2, pp. 255--335, 2020.

\bibitem{NCMEC22}
\BIBentryALTinterwordspacing
{National Center for Missing and Exploited Children}, ``{CyberTipline 2022 Report},'' 2022, accessed on 23 Jun 2023. [Online]. Available: \url{https://www.missingkids.org/gethelpnow/cybertipline/cybertiplinedata}
\BIBentrySTDinterwordspacing

\bibitem{egan2011perverted}
V.~Egan, J.~Hoskinson, and D.~Shewan, ``Perverted justice: A content analysis of the language used by offenders detected attempting to solicit children for sex,'' \emph{Antisocial behavior: Causes, correlations and treatments}, vol.~20, no.~3, p. 273297, 2011.

\bibitem{kontostathis2009chatcoder}
A.~Kontostathis, L.~Edwards, and A.~Leatherman, ``Chatcoder: Toward the tracking and categorization of internet predators,'' in \emph{Proceedings of the 7th text mining workshop}, 2009, pp. 1--7.

\bibitem{olson2007entrapping}
L.~N. Olson, J.~L. Daggs, B.~L. Ellevold, and T.~K. Rogers, ``Entrapping the innocent: Toward a theory of child sexual predators’ luring communication,'' \emph{Communication Theory}, vol.~17, no.~3, pp. 231--251, 2007.

\bibitem{leatherman2009luring}
A.~Leatherman, ``Luring language and virtual victims: Coding cyber-predators online communicative behavior,'' \emph{Media and Communication Studies, Ursinus College, Collegeville, PA}, 2009.

\bibitem{inches2012overview}
G.~Inches and F.~Crestani, ``Overview of the international sexual predator identification competition at pan-2012.'' in \emph{CLEF (Online working notes/labs/workshop)}, vol.~30.\hskip 1em plus 0.5em minus 0.4em\relax Citeseer, 2012.

\bibitem{kontostathis2012identifying}
A.~Kontostathis, A.~Garron, K.~Reynolds, W.~West, and L.~Edwards, ``Identifying predators using chatcoder 2.0.'' in \emph{CLEF (Online Working Notes/Labs/Workshop)}, 2012.

\bibitem{vogt2021early}
M.~Vogt, U.~Leser, and A.~Akbik, ``Early detection of sexual predators in chats,'' in \emph{Proceedings of the 59th Annual Meeting of the Association for Computational Linguistics and the 11th International Joint Conference on Natural Language Processing (Volume 1: Long Papers)}, 2021, pp. 4985--4999.

\bibitem{borj2021detecting}
P.~R. Borj, K.~Raja, and P.~Bours, ``Detecting sexual predatory chats by perturbed data and balanced ensembles,'' in \emph{2021 International Conference of the Biometrics Special Interest Group (BIOSIG)}.\hskip 1em plus 0.5em minus 0.4em\relax IEEE, 2021, pp. 1--5.

\bibitem{rezaee2023detecting}
P.~Rezaee~Borj, K.~Raja, and P.~Bours, ``Detecting online grooming by simple contrastive chat embeddings,'' in \emph{Proceedings of the 9th ACM International Workshop on Security and Privacy Analytics}, 2023, pp. 57--65.

\bibitem{borj2022online}
P.~R. Borj, K.~Raja, and P.~Bours, ``Online grooming detection: A comprehensive survey of child exploitation in chat logs,'' \emph{Knowledge-Based Systems}, p. 110039, 2022.

\bibitem{demant2019drug}
J.~Demant, S.~A. Bakken, A.~Oksanen, and H.~Gunnlaugsson, ``Drug dealing on facebook, snapchat and instagram: A qualitative analysis of novel drug markets in the nordic countries,'' \emph{Drug and alcohol review}, vol.~38, no.~4, pp. 377--385, 2019.

\bibitem{yang2017tracking}
X.~Yang and J.~Luo, ``Tracking illicit drug dealing and abuse on instagram using multimodal analysis,'' \emph{ACM Transactions on Intelligent Systems and Technology (TIST)}, vol.~8, no.~4, pp. 1--15, 2017.

\bibitem{mackey2018solution}
T.~Mackey, J.~Kalyanam, J.~Klugman, E.~Kuzmenko, and R.~Gupta, ``Solution to detect, classify, and report illicit online marketing and sales of controlled substances via twitter: using machine learning and web forensics to combat digital opioid access,'' \emph{Journal of medical Internet research}, vol.~20, no.~4, p. e10029, 2018.

\bibitem{li2019machine}
J.~Li, Q.~Xu, N.~Shah, and T.~K. Mackey, ``A machine learning approach for the detection and characterization of illicit drug dealers on instagram: model evaluation study,'' \emph{Journal of medical Internet research}, vol.~21, no.~6, p. e13803, 2019.

\bibitem{shah2022unsupervised}
N.~Shah, J.~Li, and T.~K. Mackey, ``An unsupervised machine learning approach for the detection and characterization of illicit drug-dealing comments and interactions on instagram,'' \emph{Substance abuse}, vol.~43, no.~1, pp. 273--277, 2022.

\bibitem{qian2021distilling}
Y.~Qian, Y.~Zhang, Y.~Ye, C.~Zhang \emph{et~al.}, ``Distilling meta knowledge on heterogeneous graph for illicit drug trafficker detection on social media,'' \emph{Advances in Neural Information Processing Systems}, vol.~34, pp. 26\,911--26\,923, 2021.

\bibitem{hu2021identifying}
C.~Hu, M.~Yin, B.~Liu, X.~Li, and Y.~Ye, ``Identifying illicit drug dealers on instagram with large-scale multimodal data fusion,'' \emph{ACM Transactions on Intelligent Systems and Technology (TIST)}, vol.~12, no.~5, pp. 1--23, 2021.

\bibitem{siddiqi2015keyword}
S.~Siddiqi and A.~Sharan, ``Keyword and keyphrase extraction techniques: a literature review,'' \emph{International Journal of Computer Applications}, vol. 109, no.~2, 2015.

\bibitem{ramos2003using}
J.~Ramos \emph{et~al.}, ``Using tf-idf to determine word relevance in document queries,'' in \emph{Proceedings of the first instructional conference on machine learning}, vol. 242, no.~1.\hskip 1em plus 0.5em minus 0.4em\relax Citeseer, 2003, pp. 29--48.

\bibitem{rose2010automatic}
S.~Rose, D.~Engel, N.~Cramer, and W.~Cowley, ``Automatic keyword extraction from individual documents,'' \emph{Text mining: applications and theory}, pp. 1--20, 2010.

\bibitem{mihalcea2004textrank}
R.~Mihalcea and P.~Tarau, ``Textrank: Bringing order into text,'' in \emph{Proceedings of the 2004 conference on empirical methods in natural language processing}, 2004, pp. 404--411.

\bibitem{grootendorst2020keybert}
\BIBentryALTinterwordspacing
M.~Grootendorst, ``Keybert: Minimal keyword extraction with bert.'' 2020. [Online]. Available: \url{https://doi.org/10.5281/zenodo.4461265}
\BIBentrySTDinterwordspacing

\bibitem{devlin2018bert}
J.~Devlin, M.-W. Chang, K.~Lee, and K.~Toutanova, ``{BERT: Pre-training of deep bidirectional transformers for language understanding},'' \emph{arXiv preprint arXiv:1810.04805}, 2018.

\bibitem{sun2021capturing}
S.~Sun, Z.~Liu, C.~Xiong, Z.~Liu, and J.~Bao, ``Capturing global informativeness in open domain keyphrase extraction,'' in \emph{Natural Language Processing and Chinese Computing: 10th CCF International Conference, NLPCC 2021, Qingdao, China, October 13--17, 2021, Proceedings, Part II 10}.\hskip 1em plus 0.5em minus 0.4em\relax Springer, 2021, pp. 275--287.

\bibitem{kontostathis2010text}
A.~Kontostathis, L.~Edwards, and A.~Leatherman, ``Text mining and cybercrime,'' \emph{Text mining: Applications and theory}, pp. 149--164, 2010.

\bibitem{souza2020bertimbau}
F.~Souza, R.~Nogueira, and R.~Lotufo, ``Bertimbau: pretrained bert models for brazilian portuguese,'' in \emph{Intelligent Systems: 9th Brazilian Conference, BRACIS 2020, Rio Grande, Brazil, October 20--23, 2020, Proceedings, Part I 9}.\hskip 1em plus 0.5em minus 0.4em\relax Springer, 2020, pp. 403--417.

\bibitem{meng2017deep}
R.~Meng, S.~Zhao, S.~Han, D.~He, P.~Brusilovsky, and Y.~Chi, ``Deep keyphrase generation,'' \emph{arXiv preprint arXiv:1704.06879}, 2017.

\bibitem{xiong2019open}
L.~Xiong, C.~Hu, C.~Xiong, D.~Campos, and A.~Overwijk, ``Open domain web keyphrase extraction beyond language modeling,'' \emph{arXiv preprint arXiv:1911.02671}, 2019.

\end{thebibliography}
\balance

\end{document}